\documentclass[runningheads]{llncs}

\usepackage{eccv}

\usepackage{eccvabbrv}

\usepackage{graphicx}
\usepackage{booktabs}

\usepackage[accsupp]{axessibility}  

\usepackage[pagebackref,breaklinks,colorlinks,citecolor=eccvblue]{hyperref}

\usepackage{orcidlink}

\usepackage{multirow} 

\usepackage{enumitem}

\begin{document}

\title{GaussReg: Fast 3D Registration with Gaussian Splatting} 

\titlerunning{GaussReg}

\renewcommand{\thefootnote}{\fnsymbol{footnote}}
\footnotetext[4]{\label{footnote:corr}Corresponding author.}
\author{Jiahao Chang\inst{1} \and Yinglin Xu\inst{2} \and Yihao Li\inst{2} \and Yuantao Chen\inst{1} \and Xiaoguang Han\inst{1,2}$^{\ref{footnote:corr}}$}


\authorrunning{Chang et al.}

\institute{School of Science and Engineering, The Chinese University of Hong Kong, Shenzhen \and
The Future Network of Intelligence Institute, CUHK-Shenzhen \\
\url{https://jiahao620.github.io/gaussreg}
}

\maketitle

\begin{abstract}

    Point cloud registration is a fundamental problem for large-scale 3D scene scanning and reconstruction. 
    With the help of deep learning, registration methods have evolved significantly, reaching a nearly-mature stage. 
    As the introduction of Neural Radiance Fields (NeRF), it has become the most popular 3D scene representation as its powerful view synthesis capabilities.
    Regarding NeRF representation, its registration is also required for large-scale scene reconstruction.
    However, this topic extremly lacks exploration. 
    This is due to the inherent challenge to model the geometric relationship among two scenes with implicit representations.
    The existing methods usually convert the implicit representation to explicit representation for further registration.
    Most recently, Gaussian Splatting (GS) is introduced, employing explicit 3D Gaussian. 
    This method significantly enhances rendering speed while maintaining high rendering quality.
    Given two scenes with explicit GS representations, in this work, we explore the 3D registration task between them. 
    To this end, we propose GaussReg, a novel coarse-to-fine framework, both fast and accurate. 
    The coarse stage follows existing point cloud registration methods and estimates a rough alignment for point clouds from GS.
    We further newly present an image-guided fine registration approach, which renders images from GS to provide more detailed geometric information for precise alignment. 
    To support comprehensive evaluation, we carefully build a scene-level dataset called ScanNet-GSReg with $1379$ scenes obtained from the ScanNet dataset and collect an in-the-wild dataset called GSReg.
    Experimental results demonstrate our method achieves state-of-the-art performance on multiple datasets.
    Our GaussReg is $44 \times$ faster than HLoc (SuperPoint as the feature extractor and SuperGlue as the matcher) with comparable accuracy.
\end{abstract}

\section{Introduction}
\label{sec:intro}

\begin{figure}[ht]
    \includegraphics[width=1\linewidth]{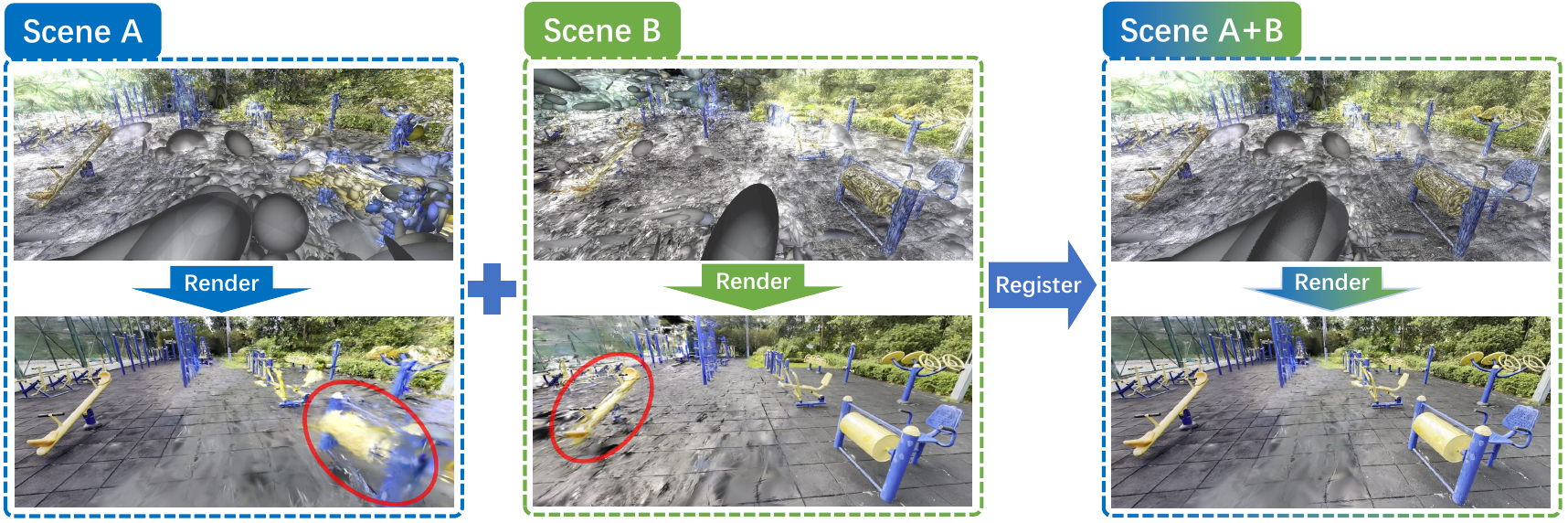}
    \caption{
    The purpose of our method is to register scenes A and B with Gaussian Splatting\cite{kerbl3Dgaussians} models, and then combine A with B to get the fused Gaussian Splatting model.
    The first row is the visualization of the 3D Gaussians.
    }
    \label{fig:method}
    \end{figure}

In traditional 3D scene scanning and reconstruction, a large-scale scene is usually divided into different blocks, resulting in many independent sub-scenes that may not in the same coordinate system. 
Therefore, the registration between them plays a crucial role.
Currently, point cloud registration has been widely studied and reached a relative mature stage, with several representative works such as ICP~\cite{121791}, D3Feat~\cite{Bai_2020_CVPR}, Geotransformer~\cite{9879611}, etc.
The mainstream methods typically involve extracting features from point clouds and  locating matching points to calculate the transformation between the two input scenes.

Recently, a new 3D representation - Neural Radiance Fields (NeRF) has been introduced and quickly gained attention due to its powerful capability in view synthesis, and it has been widely used in representing 3D scenes. 
When considering large-scale scene reconstruction based on NeRF, there are two main challenges: 
1) Due to the complex occlusions present in real-world scenes, lots of images or videos are often required to capture for large-scale reconstruction, leading to a time-consuming data collection process.
2) Optimizing NeRF with numerous images is computationally intensive. Therefore, a direct approach is to divide a large-scale scene into some smaller scenes, reconstruct them separately, and then use registration to combine all these small scenes together.

Consider two overlapping scenes, each with its own NeRF model.
Currently, the methods for registering two reconstructed NeRF scenes can be generally categorized into two types: 
1) As the method proposed in NeRFuser~\cite{fang2023nerfuser}, we can render a large number of images for each scene, then recover poses of all these images together from structure-from-motion (SfM).
However, this method is very time-consuming; 
2) As in the method DReg-NeRF~\cite{Chen_2023_ICCV}, we can convert the implicit radiance field to explicit voxel by querying voxel grids from NeRF of two scenes, and extract features to establish their matching relationship for registration.
But this method faces two issues: 
a) it is difficult to turn NeRF of unbounded scene to bounded voxel;
b) the resolution limitation of the voxel grid makes this method unsuitable for larger scenes.

Most recently, Gaussian Splatting (GS)~\cite{kerbl3Dgaussians} has been proposed, which introduces an explicit representation of 3D Gaussians, ensuring high-quality rendering while speeding up the rendering process. 
Then, an interesting question comes up: ``\textit{As GS provides a point-like representation, can we conduct GS registration resorting to point-cloud registration methods? }''

In this work, we explore fast and accurate 3D registration with GS to answer the question. Taking GS models of the two scenes as input, we first extract their point clouds from GS.
Thus, the straightforward approach is to adapt point cloud registration methods to the registration between these GS point clouds. 
To this end, a coarse registration method is designed which follows standard point cloud registration pipeline, such as GeoTransformer ~\cite{9879611}, but with special consideration of extra attributes (e.g., opacity) in 3D gaussians. 




Compared with traditionally collected point cloud data, point clouds from GS only capture rough geometric structure and are usually noisy. 
Thus, a coarse registration can not achieve precise results with sufficient accuracy. 
We further propose a novel image-guided fine registration pipeline built upon the coarse registration result. 
Our main idea is from the observation that GS not only contains geometry information but also inherently detailed image information, which can support more accurate alignment. Therefore, we first locate the overlapping regions with the help of coarse registration, where a few images are rendered with the help of GS. 
Then, the fine registration pipeline projects images into 3D volumetric features for final matching and transformation estimation. 

Ultimately, we propose a novel coarse-to-fine GS registration framework: GaussReg. 
However, it still lacks evaluation benchmarks of scene-level registration with GS. 
To support this, we construct a dataset called ScanNet-GSReg, comprising 1379 scenes from the ScanNet~\cite{dai2017bundlefusion} dataset.
In addition, we collect a dataset named GSReg, comprising 6 indoor and 4 outdoor scenarios, to assess the generalization capability of our method.
We conduct extensive experiments on the ScanNet-GSReg dataset, the Objaverse~\cite{objaverseXL} dataset used in DReg-NeRF~\cite{Chen_2023_ICCV}, and the GSReg dataset, demonstrating the effectiveness of our method.

The main contributions can be summarized as:
\begin{itemize}[noitemsep,topsep=0pt]
    
    
    
    \item[$\bullet$] To the best of our knowledge, we are the first to explore the registration of 3D scenes considering Gaussian Splatting representations. 

    \item[$\bullet$] We carefully designed a novel coarse-to-fine pipeline that fully considers the characteristics of 3D gaussians, which performs both fast and accurate.

    \item[$\bullet$] An image-guided fine registration is newly presented that takes rendered images of GS into account for fine-level alignment. We also believe this strategy opens minds fro GS-related researches. 

    \item[$\bullet$] A benchmark is also newly built for the proposed new task, which includes scenes from ScanNet and several self-collected in-the-wild scenes. 
    
    
\end{itemize}


\section{Related Work}

\paragraph{\textbf{3D point cloud registration}}
3D point cloud registration has been developed for decades. Given two overlapping point clouds with different coordinate systems, the target of this task is to find the transformation between them.
Traditional methods ~\cite{121791, 7349220, 9318535, 8897021, 5432191, 8490968, 9336308} divide this process into two parts: correspondence searching and transformation estimation. 
Correspondence searching involves finding sparse matched feature points between the source and target point clouds. 
Transformation estimation is to calculate the transformation matrix using these correspondences.
These two stages will be conducted iteratively to find the optimal transformation. However, these methods require many complex strategies~\cite{7349220, 9318535, 8490968, 9336308} to overcome noise, outliers, or density variations. 
To overcome these problems, deep feature extractors~\cite{Wang_2019_ICCV, gojcic20193DSmoothNet, zeng20163dmatch} are proposed to find more robust correspondences between two point clouds. 
3DRegNet~\cite{9156303} goes one step further to learn transformation between point clouds end-to-end. 
Recently, REGTR~\cite{Yew_2022_CVPR} incorporate self-attention and cross-attention mechanisms and MAC~\cite{zhang20233d} utilizes graph networks to further improve the robustness of end-to-end point cloud registration. 
GeoTransformer~\cite{9879611} proposes a geometric transformer to match superpoint~\cite{DeTone_2018_CVPR_Workshops} features and utilizes an overlap-aware circle loss for better convergence. 
New approaches are constantly being proposed, proving the importance of this task in scene reconstructions.

\paragraph{\textbf{3D scene representation}}
Furukawa and Ponce~\cite{5226635} provide a comprehensive classification of 3D reconstruction methods, categorizing them into four primary scene representations: volumetric fields~\cite{10.1007/978-3-7091-6756-4_6, Paris2006}, point clouds~\cite{7780814}, 3D meshes~\cite{HernndezEsteban2004, 10.1007/978-3-540-76390-1_17}, and depth maps~\cite{10.1007/978-3-030-01237-3_47, 1641047, 1641048}. Except for these representations, NeRF~\cite{mildenhall2020nerf} introduces an innovative approach by leveraging a neural implicit field to model the scene. NeRF utilizes an MLP network to optimize a 5D function (3d position plus 2d viewing direction) from a set of training images which can be used to implicitly model the scene. It has shown impressive results in image reconstruction and novel view synthesis and is widely recognized as the first photorealistic 3D scene reconstruction method. 
Various types of NeRFs have been proposed for acceleration~\cite{mueller2022instant, Chen2022ECCV, Sun_2022_CVPR} and better rendering quality~\cite{9878829, 9710056, wang2023f2nerf}. 
Another recent advancement, 3D Gaussian splatting~\cite{kerbl3Dgaussians} utilizes explicit 3D Gaussians to represent the scene. 
Each Gaussian is characterized by a covariance matrix, a center point, and opacity for a flexible optimization regime. 
The model's efficient differentiable rasterization implementation and well-designed architecture enable rapid training and real-time rendering.
Moreover, The optimization strategy is cleverly designed to adaptively control the Gaussians for ensuring very high rendering quality. 
Despite the fast innovation of scene representations, 3D registration remains to be an important issue for stable large-scale reconstructions, thus developing new registration methods for different representations is crucial.


\paragraph{\textbf{NeRF Registration}}
Neural Implicit Field~\cite{mildenhall2020nerf} has been widely accepted as a new scene representation, several methods have been proposed to do NeRF registration. 
NeRF2NeRF~\cite{goli2023nerf2nerf} utilizes human-annotated key points to obtain an initial transformation and refines it using a surface field distilled from a pre-trained NeRF. 
DReg-NeRF~\cite{Chen_2023_ICCV} extracts features from the occupancy grid of NeRF and applies a decoupling model~\cite{Yew_2022_CVPR} for NeRF registration, eliminating the need for human interaction in the registration process. However, it's hard to generalize to larger scenes due to its global feature-extracting strategy. NeRFuser~\cite{fang2023nerfuser} directly uses the structure from motion method to estimate the transformation using rendered images from NeRF which is very time-consuming. 
CL-NeRF~\cite{wu2023clnerf} concentrates on the continual learning of NeRF models and proposes an expert adaptor for learning newly changed scenes without finetuning the whole network. 
Most recently, 3D Gaussian splatting has been proposed as a promising scene representation, to the best of our knowledge, we are the first to propose registration methods for 3D Gaussian Splatting and achieve SOTA performance with faster registration speed and better rendering quality.
Moreover, continual learning and modifying scenes can be naturally done using our pipeline.

\section{Method}
In this section, we present our proposed GaussReg for 3D Registration with Gaussian Splatting (GS).
The overall architecture is illustrated in Figure~\ref{fig:method}. 

\begin{figure}[ht]
\centering
\includegraphics[width=1.\linewidth]{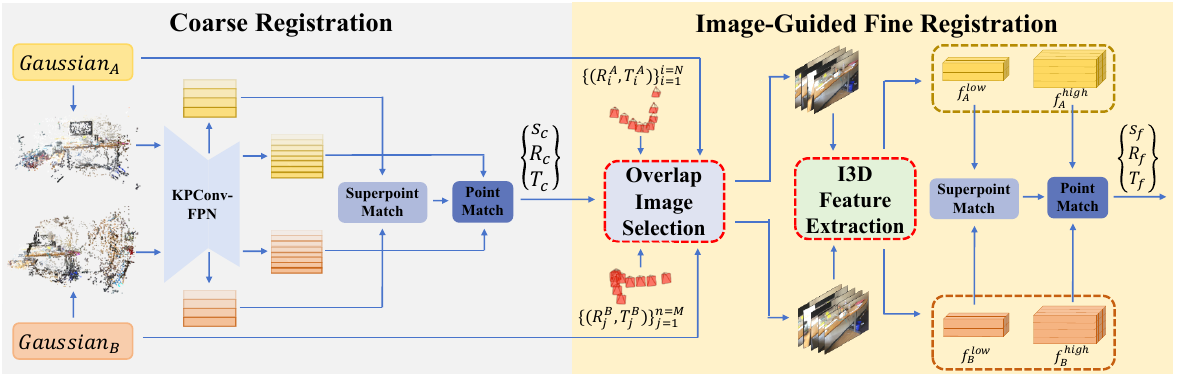}
\caption{
The architecture of GaussReg. 
Please refer to the text for detailed architecture. 
}
\label{fig:method}
\end{figure}

\subsection{Overview}\label{3.1}
As shown in Figure~\ref{fig:method}, the proposed GaussReg mainly consists of two stages, including the Coarse Registration, and the Image-Guided Fine Registration.
Here we give a brief introduction to the entire process. 
Assuming two overlapping scenes A and B, each with its own GS model,
only the camera poses of all training images are saved and accessible. 
We denote the camera poses of all training images as $\{C^A_i = (R^A_i,  T^A_i)\}^N_{i=0}$ and $\{C^B_j=(R^B_j, T^B_j)\}^M_{j=0}$ for A and B respectively. 
The GS models are denoted as $Gaussian_{A}$ and $Gaussian_{B}$ and the derived point clouds from GS models are termed as $Points_{A}$ and $Points_{B}$. 
Our goal is to discover the rigid transformation $\{s, R, T\}$ that makes scene B align with A, where $s \in \mathbb{R}$ means scale factor, $R \in \mathbb{R}^{3 \times 3}$ means rotation matrix, and $T \in \mathbb{R}^3$ represents translation vector. The coarse registration directly accepts $Points_{A}$ and $Points_{B}$ as input, and output a coarse transformation $\{s_c, R_c, T_c\}$. 
Since the extracted point cloud from a GS model tends to be noisy and distorted, the coarse alignment often needs to be more accurate.
Then, in the image-guided fine registration, we first locate a highly overlapping region based on the coarse alignment result. Around the highly overlapping region, two subsets of cameras are then selected from $\{C^A_i\}$ and $\{C^B_j\}$ respectively, from which we render several images. After that, an Image-Guided 3D (I3D) Feature Extraction is adopted to obtain volumetric features from images, which are used for subsequent local matching, ultimately achieving the accurate transformation output $\{s_f, R_f, T_f\}$.

\subsection{Coarse Registration}\label{3.2}
As we all know, the GS model is stored in the form of 3D gaussians. Each 3D gaussian stores the position $(x,y,z)$, opacity $\alpha$, rotation, scale, and the coefficients of the spherical harmonics. First, we select those confident points with the opacity $\alpha$ greater than a threshold (0.7 is chosen empirically). 
For each sampled point, the color $(r,g,b)$ is determined via spherical harmonic functions.
Finally, for every point in $Points_{A}$ or $Points_{B}$, we use $(x,y,z,\alpha,r,g,b)$ as the input channel to feed into the coarse registration pipeline.  


As shown in Figure~\ref{fig:method}, the Coarse Registration follows the workflow as GeoTransformer~\cite{9879611}, we extract multi-scale features of each point cloud through a shared KPConv-FPN~\cite{thomas2019KPConv}.
The coarsest level point features $F^{low}_{A}$ and $F^{low}_{B}$ are used for Superpoint Match and the finest level point features $F^{high}_{A}$ and $F^{high}_{B}$ are used for Point Match.
The process of Superpoint Match refers to Geotransformer~\cite{9879611}.
Noted that in Point Match, we directly utilize the ICP~\cite{121791} algorithm to obtain the coarse registration results between GS, instead of the Local-to-Global Registration in Geotransformer~\cite{9879611}.

\paragraph{\textbf{Training Strategy and Loss Function}}
Due to the scale uncertainty in monocular video reconstruction, we performed data augmentation not only on rotation and translation but also on scaling for the input Gaussian point cloud.
Even though we normalized the scale of input point clouds within a certain range, such data augmentation still preserves the diversity of relative scale differences between the point clouds to be matched.

We apply two loss functions (overlap-aware circle loss and point matching loss) from the GeoTransformer~\cite{9879611} to constrain our coarse registration network.

\begin{figure}[t]
\centering
\includegraphics[width=1.\linewidth]{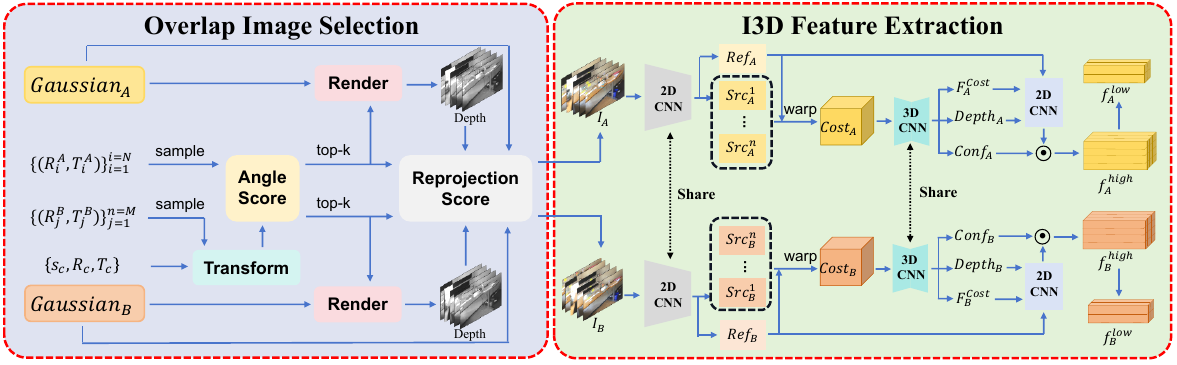}
\caption{
The illustration of our overlap image selection and I3D feature extraction. 
}
\label{fig:fine_registration}
\vspace{-5mm}
\end{figure}

\subsection{Image-Guided Fine Registration}\label{3.3}
Since the GS model doesn't impose specific geometric constraints during training, resulting point clouds may exhibit some degree of distortion. 
Relying solely on GS models might not guarantee accurate registration results. 
Considering that GS inherently contain detailed image information, an image-guided fine registration is proposed. Our key idea is to first locate overlapping regions between scene $A$ and scene $B$ and render some training images covering the region to support more precise geometric features for fine alignment. Specifically, as shown in Figure~\ref{fig:method}, our Image-Guided Fine Registration primarily involves two steps: 
1) Efficiently and accurately selecting highly overlapping cameras and rendering images accordingly;
2) Utilizing these images to construct volumetric features for further fine registration.

\paragraph{\textbf{Overlap Image Selection}}
As shown in Figure~\ref{fig:fine_registration}, the main goal of this part is to find two small subsets of cameras, from $\{C^A_i\}$ and $\{C^B_j\}$ respectively, which share as large common perspective area as possible.  
Before selection, we first uniformly sampled two subsets $\{C^a_i\}$ and $\{C^b_j\}$ to reduce computational cost, and then apply $\{s_c, R_c, T_c\}$ on $\{C^b_i\}$ for a coarse alignment, causing $\{\hat{C}^b_i\}$. 
Each subset contains $30$ images in our experiments.
Our selection follows 3 steps: 1) For every pair $(C^a_i, \hat{C}^b_j)$, we calculate the cosine value of the angle between their camera orientations. Finally, top-k closest pairs will be kept, where $k=10$ in our experiments. 
Thanks to the coarse alignment, this step can accurately and quickly removes many useless pairs; 
2) To achieve more accurate selection, for each pair $(C^a_p, \hat{C}^b_q)$  remained after step 1, we further calculate the area of their perspective sharing. To do so, two low-resolution depth maps $d^a_p$ and $d^b_q$ are rendered from $Gaussian_{A}$ and $Gaussian_{B}$ respectively. Then, we calculate how portion of points derived from $d^a_p$ can be seen from $\hat{C}^b_q$ and how portion of points derived from $d^b_q$ can be seen from $C^a_p$. With the evaluation of the averaged portion, we find the closest pair $(C^a_{i0}, \hat{C}^b_{j0})$. 
Thanks to the fast rendering speed of GS, the depth map rendering is done efficiently; 
3) We finally pick two subsets of training cameras respectively in the neighborhood of $C^a_{i0}$ and $\hat{C}^b_{j0}$. 
Under the selected cameras, the image sets $I_A$ and $I_B$ are obtained by rendering from $Gaussian_{A}$ and $Gaussian_{B}$ to be fed into the next feature extraction stage.

\paragraph{\textbf{Image-Guided 3D Feature Extraction}}

As shown in Figure~\ref{fig:fine_registration}, we adopt the principle of multi-view stereo (MVS) by utilizing images to assist in estimating the depth and extracting volumetric features of the reference image.
Without loss of generality, we use scene $A$ as an example in the following description.
First of all, we input $I_A$ into a 2D convolutional neural network to get features $Ref_A, \{Src^k_A\}^n_{k=0}$, which turn into the cost volume $Cost_A$ according to the depth hypotheses $\{d_{l}\}^D_{l=0}$ by differentiable homography.
Building the cost volume requires the minimum and maximum distances, which can be automatically computed from the rendered depth map of the reference image.
Followed by the 3DCNN regularization, the probability volume $P_A \in R^{D \times H \times W}$ and feature volume $F_A \in R^{C \times D \times H \times W}$ are obtained from the cost volumes, where $C$ is the number of feature channels, and $(H, W)$ is the resolution of $Ref_A$.
For any pixel $p$ on $Ref_A$, our network predicts a probability distribution $\{P_A^l(p)\}^D_{l=0}$.
We pick out $l_{0}$ satisfying:
\begin{equation}\label{eq:eq3.2}
\noindent l_{0} = \mathop{\arg\max}_{l=0,1,...,D-1}\{P_A^{l}(p) + P_A^{l + 1}(p)\},
\end{equation}
where $P_A^{l}(p)$ represents the probability of the pixel $p$ being at depth $d_l$.
%
The feature from cost volume $F^{Cost}_A(p)$, predicted depth $Depth_A(p)$ and confidence map $Conf_A(p)$ are calculated by:
\begin{small}
\begin{equation}\label{eq:eq3.3}
\begin{aligned}
&F^{Cost}_A(p) = F^{l_0}_A(p) \cdot \frac{P_A^{l_0}(p)}{P_A^{l_0}(p) + P_A^{l_0 + 1}(p)} + F^{l_0}_A(p) \cdot \frac{P_A^{l_0+1}(p)}{P_A^{l_0}(p) + P_A^{l_0 + 1}(p)},
\\
&Depth_A(p) = d_{l_0} \cdot \frac{P_A^{l_0}(p)}{P_A^{l_0}(p) + P_A^{l_0 + 1}(p)} + d_{l_0 +1} \cdot \frac{P_A^{l_0+1}(p)}{P_A^{l_0}(p) + P_A^{l_0 + 1}(p)},
\\
&Conf_A(p) = P_A^{l}(p) + P_A^{l + 1}(p).
\end{aligned}
\end{equation}
\end{small}
%
Then we concatenate $Ref_A$, $F^{Cost}_A$, and $Depth_A$, and pass them through convolutional layers.
After confidence-based filtering, we obtain high-resolution feature $f^{high}_A$ and low-resolution feature $f^{low}_A$.
This process can be described as

\begin{small}
\begin{equation}\label{eq:eq3.4}
\begin{aligned}
&f^{high}_A = Conv(Concate(Ref_A, F^{Cost}_A, Depth_A))[Conf_A > Mean(Conf_A)],
\\
&f^{low}_A = Conv(f^{high}_A)[Conf_A > Mean(Conf_A)],
\end{aligned}
\end{equation}
\end{small}
$f^{high}_B$ and $f^{low}_B$ are obtained in the same manner.
Next, we project the features into the coordinate system of $Gaussian_{A}$ according to the corresponding depth maps.
Finally, following the same procedure as in coarse registration, we obtain the fine registration result $\{s_f, R_f, T_f\}$.

\paragraph{\textbf{Training Strategy and Loss Function}}

Overlap Image Selection is not involved in the training of the fine registration network. 
We randomly sample pairs of multi-view images with overlap from ScanNet dataset for training. 
During training, we also apply data augmentation to the camera extrinsic.

Our loss function mainly consists of two parts, depth loss and registration loss.
Depth loss is a cross-entropy loss to supervise the probability volume:
\begin{equation}\label{eq:eq3.5}
\noindent L_{depth} = \sum_{p\in\Omega_A} -P^{gt}_A(p)logP_A(p) + \sum_{p\in\Omega_B} -P^{gt}_B(p)logP_B(p),
\end{equation}
where $\Omega_A$ and $\Omega_B$ are the sets of valid points.
$P^{gt}_A(p)$ and $P^{gt}_B(p)$ denote the one-hot labels from the ground-truth depth of $p$. 
$P_A(p)$ and $P_B(p)$ denote the predicted probability distribution of $p$.
Registration loss $L_{regis}$ is the same as loss function used in coarse registration.
Therefore, our total loss in the fine registration network is:
\begin{equation}\label{eq:eq3.6}
\noindent L_{total} = \lambda L_{depth} + L_{regis},
\end{equation}
where $\lambda = 10$ in our experiments.

\begin{figure}[t]
\centering
\includegraphics[width=1\linewidth]{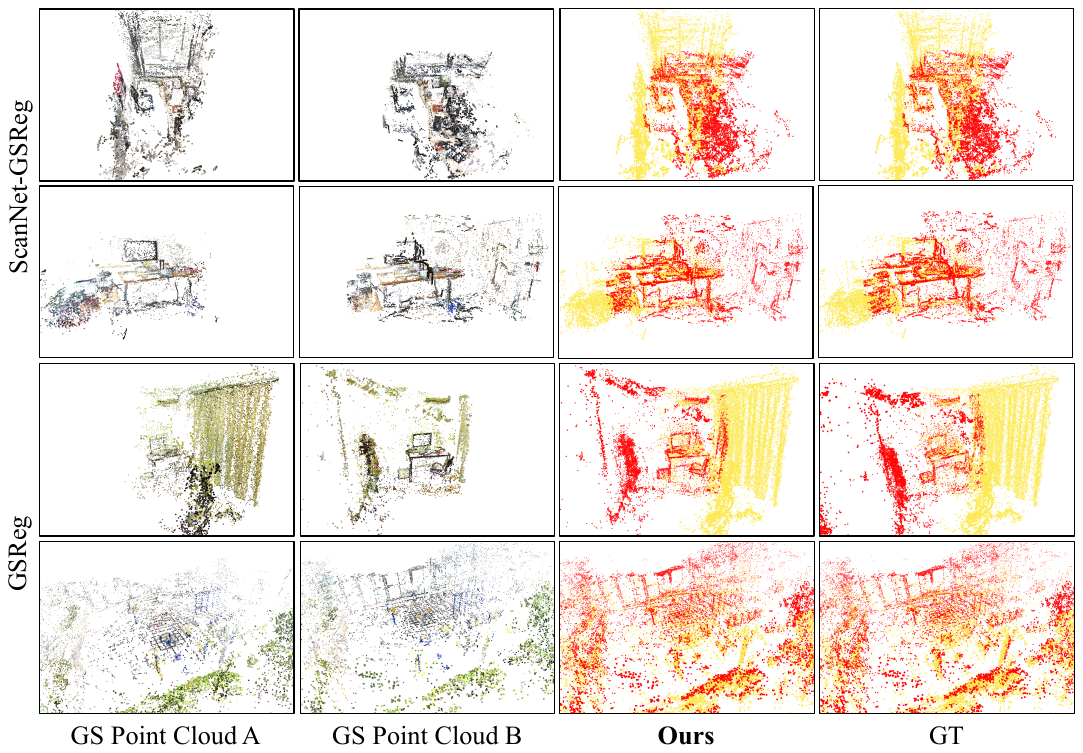}
\caption{
Visualization of our final registration results on ScanNet-GSReg and GSReg.
The first two columns are visualizations of GS point clouds to be registered.
The last two columns are visualizations of our final registration and ground-truth results.
}
\label{fig:visualization}
\end{figure}

\subsection{Gaussian Splatting Fusion and Filtering}\label{3.4}

After obtaining the final registration result, it is time to merge the two GS models. 
To transform $Gaussian_B$ into the coordinate system of $Gaussian_A$, denoted as $Gaussian_{B \rightarrow A}$, we start by transforming the position of the 3D gaussian:
\begin{equation}\label{eq:eq4.1}
\noindent (x_{B \rightarrow A}, y_{B \rightarrow A}, z_{B \rightarrow A})^T = s_f R_f (x_{B}, y_{B}, z_{B})^T + T_f.
\end{equation}
The opacity is invariant to the transformation $\alpha_{B \rightarrow A} = \alpha_{B}$.
The rotation $R_{B \rightarrow A} \in R^{3 \times 3}$ and scale $S_{B \rightarrow A} \in R^{3}$ of the 3D gaussian can be computed as:
\begin{equation}\label{eq:eq4.2}
\begin{aligned}
&R_{B \rightarrow A} = R_f R_{B},
\\
&S_{B \rightarrow A} = s_f S_{B}.
\end{aligned}
\end{equation}
From the properties of spherical harmonics (SH) coefficients, we know that the rotation of SH coefficients is a linear transformation of the SH coefficients, and the rotation of each order of SH coefficients can be performed separately.
Hence, for the $i$-th order of SH coefficients, we can obtain the transformation of SH coefficients through the following approach:
1) Select any 2i+1 unit vectors $u_0, ..., u_{2i+1}$, let $Q = (SH(u_0), ..., SH(u_{2i+1}))$, where $SH$ is the function that projects the direction vector to the corresponding SH values;
2) Apply transformation $\{s_f, R_f, T_f\}$ to vectors $u_0, ..., u_{2i+1}$ to yield $\hat{u_0}, ..., \hat{u_{2i+1}}$;
3) $(SH(\hat{u_0}), ..., SH(\hat{u_{2i+1}})) Q^{-1}$ is the transformation matrix of SH coefficients.
Note that, it is difficult to choose vectors $u_0, ..., u_{2i+1}$ to ensure that $Q$ is invertible, so in our experiments, we use the pseudo inverse as an approximation to inverse while calculating $Q^{-1}$.
Finally, we merge the 3D gaussians in $Gaussian_A$ closer to the center of $A$ with the 3D gaussians in $Gaussian_B$ closer to the center of $B$ to get $Gaussian_{A+B}$.

\section{Experiment}
\subsection{Experiment Setup}\label{3.1}

\begin{table}[t]
\centering
\caption{Evaluation on the ScanNet-GSReg dataset. $\downarrow$ means lower is better, and $\uparrow$ means higher is better.
We believe that data preprocessing is also part of the method, so we include the time of obtaining point cloud from GS in the time reported.
Here HLoc~\cite{sarlin2019coarse}* means HLoc~\cite{sarlin2019coarse} (SuperPoint~\cite{DeTone_2018_CVPR_Workshops} + SuperGlue~\cite{sarlin20superglue}).
}
\label{tab:table1}
\scalebox{0.95}{
\begin{tabular}{c|ccccc}
\hline
Methods                & RRE$\downarrow$ & RTE$\downarrow$ & RSE$\downarrow$ & Succss Ratio$\uparrow$ & Time(s)$\downarrow$ \\ \hline
HLoc~\cite{sarlin2019coarse}* & \textbf{2.725}        & 0.099        & 0.098        & 0.756                  & 212.3            \\ \hline
FGR~\cite{10.1007/978-3-319-46475-6_47} & 157.126  & 3.328  & 0.268        & \textbf{1.000}                  & \textbf{3.4}           \\ 
\hline
REGTR~\cite{Yew_2022_CVPR} & 80.095  & 2.768  & 0.408        & \textbf{1.000}                  & 3.5           \\ 
\hline
\textbf{Ours}                  & 2.827        & \textbf{0.042}        & \textbf{0.032}        & \textbf{1.000}                     & 4.8              \\ \hline
\end{tabular}
}
\end{table}

\paragraph{\textbf{Dataset}}

As there is currently no scene-level dataset available for our task, it is necessary for us to create a dataset in order to evaluate GS registration.
ScanNet~\cite{dai2017bundlefusion} is a frequently used 3D dataset for indoor scenes, consisting of $1513$ training scenes and $100$ test scenes. 
Each scene in ScanNet includes camera intrinsics, a sequence of images, along with the corresponding camera extrinsics and depth maps. 
Therefore, we decide to build a dataset based on ScanNet, called \textbf{ScanNet-GSReg dataset}.
First, we randomly sample two continue image sequences from each scene. 
Each sequence contains $80$ to $120$ images, and the sampling interval ranges from  $1$ to $5$.
The overlap ratio, calculated as the proportion of repeated images between two sequences, ranges from $0.2$ to $0.8$. 
Then, we apply random transformations to each set of camera extrinsics independently to simulate the inconsistency between the world coordinates of the two sequences and record these two transformations as the ground-truth transformation between their world coordinates.
Using these image sequences and corresponding camera parameters, we reconstruct the GS models separately.
Each model undergoes $10000$ iterations of training.
Eventually, after excluding cases of failed initial point cloud generation or unsuccessful GS reconstruction, we obtain $1297$ training samples and $82$ test samples. 
Furthermore, to validate the generalization of our method, we collected 10 real-world scenes for testing, called \textbf{GSReg dataset}, which includes $6$ indoor and $4$ outdoor scenes. 
For each scene, we record two videos. 
First, we use HLoc~\cite{sarlin2019coarse} (SuperPoint~\cite{DeTone_2018_CVPR_Workshops} as the feature extractor and SuperGlue~\cite{sarlin20superglue} as the matcher) to obtain the camera poses individually for each video, and then combine the two videos for a joint camera pose estimation to obtain the ground-truth transformation between the two GS models. 
To evaluate the performance of GaussReg on objects, we also conduct tests on the \textbf{Objaverse dataset}~\cite{objaverseXL} used in DReg-NeRF~\cite{Chen_2023_ICCV}, whose test set contains 44 objects.


\paragraph{\textbf{Metric}}

We refer to metrics of point cloud registration as in~\cite{9879611} and modify them to account for scale factors.
Finally, we evaluate GaussReg on the ScanNet-GSReg and GSReg datasets with three metrics: 
1) Relative Rotational Error (RRE), the geodesic distance between the estimated and ground-truth rotation matrix; 
2) Relative Translation Error (RTE), the ratio of the Euclidean distance between the estimated and ground-truth translation vectors to the norm of the ground-truth translation vector; 
3) Relative Scale Error (RSE), the ratio of the Euclidean distance between the estimated and ground-truth scale factors to the ground-truth scale factor.
For a fair comparison, we follow DReg-NeRF~\cite{Chen_2023_ICCV} to evaluate GaussReg on the Objaverse dataset with two metrics: 
1) Relative Rotational Error (RRE); 
2) Absolute Translational Error (ATE), the Euclidean distance between the estimated and ground-truth translation vectors.

\paragraph{\textbf{Implementation Details}}

Our GaussReg is merely trained on the ScanNet-GSReg training set and evaluated on the ScanNet-GSReg test set, Objaverse test set, and GSReg dataset. 
Our method was implemented with PyTorch~\cite{paszke2017automatic}.
In the coarse registration network, we limit the number of input points to $30000$ during training. 
In the image-guided fine registration network, we render $n=5$ images per GS model as input and set the number of depth hypotheses to $D=64$. 
Both networks are trained separately for $40$ epochs with a batch size of $1$.
The learning rate starts from $1e-4$ and decays exponentially by $0.05$ every epoch.




\begin{table}[t]
\begin{minipage}[c]{0.48\textwidth}
        \centering
            \caption{Evaluation on the Objaverse dataset. $\downarrow$ means lower is better.}
            \label{tab:table2}
        \scalebox{1.}{
\begin{tabular}{c|ccc}
\hline
Methods         & RRE$\downarrow$       & ATE$\downarrow$      \\ \hline
FGR~\cite{10.1007/978-3-319-46475-6_47}                     & 61.59               & 13.50      \\ \hline
REGTR~\cite{Yew_2022_CVPR}                   & 113.78              & 43.31         \\ \hline
Dreg-NeRF~\cite{Chen_2023_ICCV}               & 9.67                & 3.85         \\ \hline
\textbf{Ours w/o. fine} & { \textbf{2.47}} & { \textbf{3.46}} \\ 
\hline
\end{tabular}
}
\end{minipage}
\hspace{.15in}
\renewcommand
\arraystretch{1.25}
\begin{minipage}[c]{0.45\textwidth}
\centering
    \caption{Evaluation on the GSReg dataset. $\downarrow$ means lower is better.}
    \label{tab:table3}
\scalebox{0.85}{
\begin{tabular}{c|ccc}
\hline 
Methods         & RRE$\downarrow$       & RTE$\downarrow$      & RSE$\downarrow$ \\ \hline
\textbf{Ours w/o. fine}                    & 6.904          & 0.074          & 0.051        \\ \hline
\textbf{Ours}                              & \textbf{2.989} & \textbf{0.065} & \textbf{0.047} \\ \hline
\end{tabular}
}
\end{minipage}
\vspace{-1.2em}
\end{table}

\subsection{Comparison with Other Methods}

\paragraph{\textbf{Evaluation on the ScanNet-GSReg Dataset}}

Due to the maturity of Structure from Motion (SFM) technology, a natural approach for 3D registration with GS is to render a large number of images and utilize SFM for joint registration.
Therefore, we select the current SOTA method, HLoc~\cite{sarlin2019coarse} (SuperPoint~\cite{DeTone_2018_CVPR_Workshops} + SuperGlue~\cite{sarlin20superglue}), as the baseline for comparison on ScanNet.
In the subsequent discussion, we refer to HLoc~\cite{sarlin2019coarse} (SuperPoint~\cite{DeTone_2018_CVPR_Workshops} + SuperGlue~\cite{sarlin20superglue}) as HLoc for brevity.
For the two GS models to be registered, we uniformly sample $30$ training poses each to render images, and use $60$ images in total for HLoc to estimate pose.
We can obtain the registration result of the two GS models following the procedure described in NeRFuser~\cite{fang2023nerfuser}.
We also evaluate traditional point cloud registration method Fast Global Registration (FGR)~\cite{10.1007/978-3-319-46475-6_47} and deep point cloud registration method REGTR~\cite{Yew_2022_CVPR} (retrained on 3DMatch) by inputting the point cloud from GS.
FGR and REGTR are also followed by the ICP solver with scaling to output the transformation results, and we also limit the number of input points to $30000$.
The quantitative results are shown in Table~\ref{tab:table1}, where the Success Ratio indicates the portion of successful registrations.  
As shown in Table~\ref{tab:table1}, for $82$ scenes in ScanNet-GSReg, HLoc only registers $75.6\%$ of them successfully, while our method achieves a $100\%$ success ratio. 
For indoor scenes in ScanNet-GSReg, SuperPoint~\cite{DeTone_2018_CVPR_Workshops} sometimes fails to extract effective keypoints, leading to registration failures.
Our method outperforms HLoc in RTE and RSE metrics and is comparable in RRE. Notably, our method was significantly faster than HLoc ($4.8s$ vs. $212.3s$). 
FGR and REGTR are slightly faster than our GaussReg, however, they perform much worse than ours.
We think the reason is that the point cloud from GS is much noisier than scanning data.
Visualizations of our method on the ScanNet-GSReg test set are presented in the first two rows of Figure~\ref{fig:visualization}.
More visual results can be found in Supplementary Material.
These experiments fully demonstrate the efficiency and accuracy of our method.

\paragraph{\textbf{Evaluation on the Objaverse Dataset}}

For a fair comparison on the Objaverse dataset~\cite{objaverseXL} used in DReg-NeRF, we assume there is no scale difference between the two GS models as in DReg-NeRF~\cite{Chen_2023_ICCV}. 
In addition, we do not adopt training poses, and only use our proposed coarse registration for comparison.
As shown in Table~\ref{tab:table2}, our coarse registration method (ours w/o. fine) significantly outperforms other methods without fine-tuning, demonstrating its strong generalization capability to objects.

\paragraph{\textbf{Evaluation on the GSReg Dataset}}

The ground-truth registration results of our GSReg dataset are obtained when HLoc was successful. 
As shown in Table~\ref{tab:table2}, our method achieves registration results close to HLoc without fine-tuning, proving the strong generalizability of our approach.
Moreover, our method (ours) significantly outperforms our coarse registration (ours w./o. fine), proving the effectiveness of our fine registration.
Visualizations of our method on the GSReg dataset are presented in the last two rows of Figure~\ref{fig:visualization}.


\begin{table}[t]
\centering
\caption{Ablation study of image-guided fine registration on the ScanNet-GSReg dataset.
$\downarrow$ means lower is better, and $\uparrow$ means lower is better.}
\label{tab:table4}
\scalebox{0.9}{
\begin{tabular}{c|c|ccccc}
\hline
Index & Methods                  & RRE$\downarrow$   & RTE$\downarrow$   & RSE$\downarrow$   & Succss Ratio$\uparrow$ & Time(s)$\downarrow$ \\
\hline
1              & Hloc~\cite{sarlin2019coarse}                            & 2.725          & 0.099          & 0.098          & 0.756                  & 212.3            \\
\hline
2              & \textbf{Ours w./o. fine}          & 3.403          & 0.061          & 0.034          & \textbf{1.000}            & \textbf{3.7}     \\
\hline
3              & \textbf{Ours w./o. fine + HLoc} & \textbf{1.104} & 0.186          & \textbf{0.278} & 0.512                  & 206.8            \\
\hline
4              & \textbf{Ours}                     & 2.827          & \textbf{0.042} & 0.032          & \textbf{1.000}            & 4.8              \\ 
\hline
\end{tabular}
}
\vspace{-3mm}
\end{table}

\subsection{Ablation Study}

To deeply analyze GaussReg, we conduct detailed ablation studies on the ScanNet-GSReg dataset to evaluate the effectiveness of the proposed components.
\paragraph{\textbf{Effectiveness of Image-Guided Fine Registration}}
HLoc can also utilize image information to refine the coarse registration.
Therefore, to validate the effectiveness of image-guided fine registration, we directly combine coarse registration with HLoc.
After obtaining the coarse registration result, we use overlap image selection to select two sets of multi-view images, $Images_A$ and $Images_B$, and jointly use $Images_A$ and $Images_B$ for pose estimation with HLoc. 
As shown in Table~\ref{tab:table4}, by comparing Index-2 with Index-4, we can see that the performance is improved, which demonstrates the effectiveness of our image-guided fine registration. 
Comparing Index-2 and Index-3, we find that although HLoc shows lower RRE, its success ratio is very low ($51.2\%$), whereas our fine registration not only outperforms HLoc in RTE and RSE metrics but also has a higher success ratio ($100\%$).
Meanwhile, our fine registration is faster than HLoc ($4.8s$ vs. $206.8s$).
In addition, we explore the effect of the top-k pairs of cameras we kept in overlap image selection. Hence, we vary k from 5 to 30. In Table~\ref{tab:table5}, there is almost no change in performance when k is larger than 10 and the performance drops when k is smaller than 10.
For the sake of accuracy and efficiency, we believe that 10 is enough for k.


\begin{table}[t]
\renewcommand
\arraystretch{0.9}
\begin{minipage}[c]{0.46\textwidth}
        \centering
        \caption{Ablation study with different k in overlap image selection on ScanNet-GSReg.
        $\downarrow$ means lower is better.}
    \label{tab:table5}
\scalebox{1}{
\begin{tabular}{c | *{3}{c}}
        \hline
        Top-k & RRE$\downarrow$ & RTE$\downarrow$ & RSE$\downarrow$ \\ 
        \hline
        5 & 3.677 & 0.115 & 0.079 \\ 
        \hline
        10 & 2.827 & \textbf{0.042} & 0.032 \\ 
        \hline
        20 & 2.604 & 0.063 & 0.044 \\ 
        \hline
        30 & \textbf{2.311} & 0.091 & \textbf{0.028} \\ 
        \hline
        \end{tabular}
        }
\end{minipage}
\hspace{.15in}
\renewcommand
\arraystretch{1.2}
\begin{minipage}[c]{0.48\textwidth}
\centering
\caption{Ablation study of image-guided 3D feature extraction on ScanNet-Reg.
$\downarrow$ means lower is better.}
            \label{tab:table6}
        \scalebox{0.8}{
\begin{tabular}{c|c|cccc}
\hline
Index & Method & RRE$\downarrow$   & RTE$\downarrow$   & RSE$\downarrow$   & RDE$\downarrow$   \\ \hline
5 & \textbf{Ours w/o. I3D}   & 3.169          & \textbf{0.036} & 0.061          & \textbf{0.066} \\
\hline
6 & \textbf{Ours}            & \textbf{2.827} & 0.042          & \textbf{0.032} & 0.080           \\ \hline
\end{tabular}
}
\end{minipage}
\vspace{-1.2em}
\end{table}


\paragraph{\textbf{Effectiveness of Image-Guided 3D Feature Extraction}}
Here, we also report the Relative Depth Error (RDE), which is the ratio of the Euclidean distance between the estimated and ground-truth depth to the ground-truth depth. 
As shown in Table~\ref{tab:table6}, in Index-5, we remove the image-guided 3D (I3D) feature extraction.
Instead, we use MVSNet~\cite{1641048} to calculate depth and project depth maps to obtain two point clouds, which serve as input to KPConv-FPN~\cite{thomas2019KPConv} to extract features for registration refinement. 
Comparing Index-5 and Index-6, we observe that although Index-5 has better depth estimation accuracy, the registration results are poor, proving that extracting geometric information from images complements feature descriptors extraction.

\subsection{Results of Gaussian Splatting Fusion and Filtering}
\begin{figure}[t]
\centering
\includegraphics[width=1\linewidth]{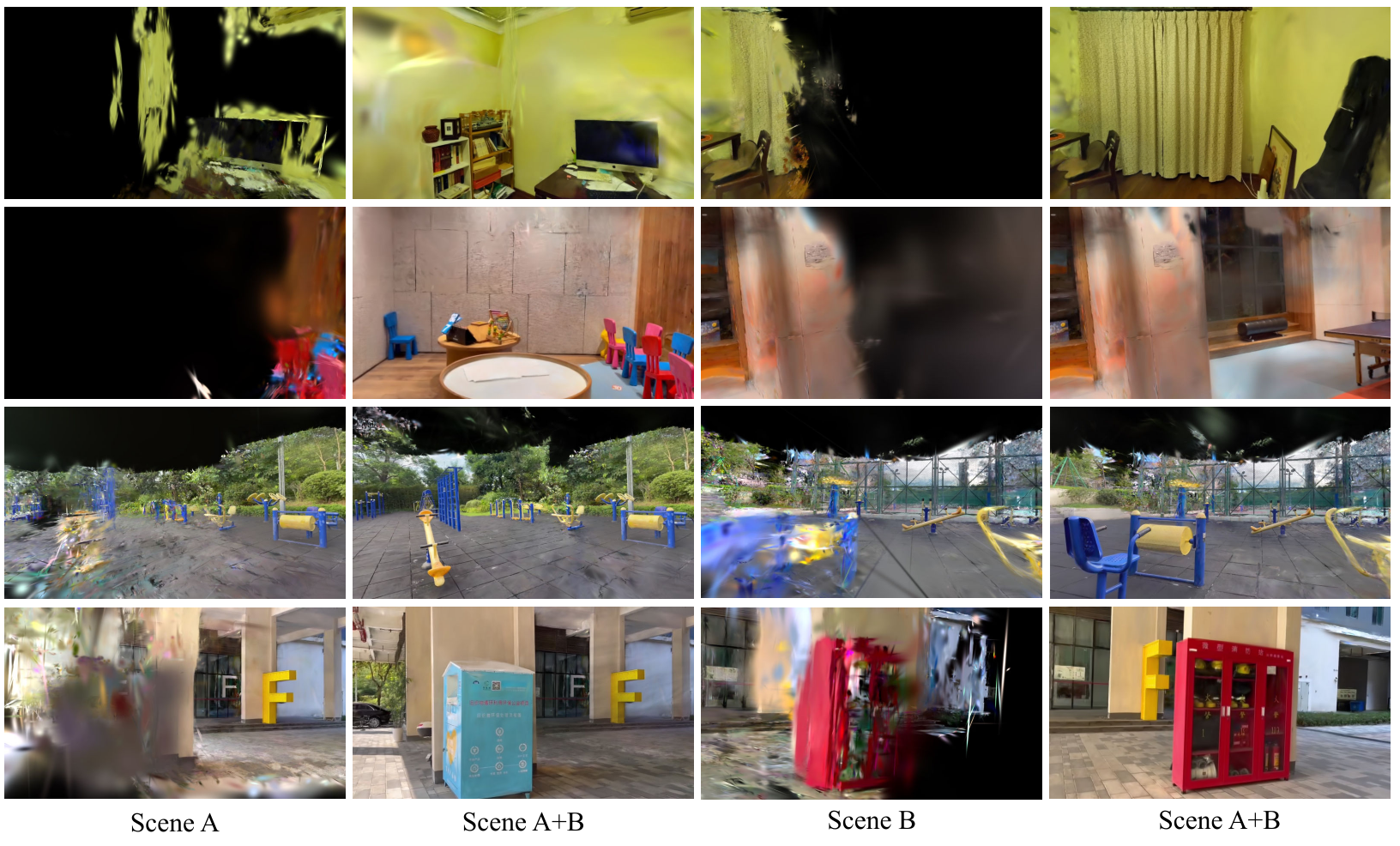}
\caption{
Quantitative Results on the GSReg dataset.
The first two rows are indoor scenes, and the last two rows are outdoor scenes.
The first and third columns are rendering images from GS~\cite{kerbl3Dgaussians} models of Scene A and Scene B.
The second and last columns are rendering images from our fused GS model.
}
\label{fig:render}
\end{figure}

In Figure~\ref{fig:render}, we present some quantitative results on GSReg dataset to demonstrate the effectiveness of our GS fusion and filtering. 
Please refer to the video attachment in Supplementary Material for the dynamic presentation. 
Our GS fusion and filtering strategy successfully merges the two GS models.
\section{Discussion}

\paragraph{\textbf{Limitations and Future Work}}

We only adopt a simple strategy to fuse and filter two GS models.
For some more complex situations, the fusion in our way is imperfect.
For instance, when two scenes are captured at different times, changes in lighting can result in differing appearances for two scenes. 
Consequently, the fused GS model obtained through our strategy may exhibit inconsistencies at the fusion boundary.
Future work can further explore to address this issue.

\paragraph{\textbf{Conclusion}}
The advent of Neural Radiance Fields (NeRF) has transformed the landscape of 3D scene representation, necessitating advancements in registration methodologies.
However, the registration of NeRF representations for large-scale scenes remains underexplored due to the inherent complexities of implicitly modeled geometric relationships. 
The recent introduction of Gaussian Splatting (GS) has significantly enhanced NeRF by introducing explicit 3D gaussians, facilitating rapid rendering while maintaining high quality.
In this study, we introduce GaussReg, a pioneering coarse-to-fine framework that utilizes GS for 3D registration with GS. 
The coarse phase leverages existing point cloud registration methods to establish a preliminary alignment for input GS point clouds. 
We innovatively devise an image-guided fine registration strategy that incorporates rendered images from these Gaussian points, enriching geometric details for accurate alignment. 
To comprehensively evaluate our approach, we construct a benchmark consisting of scenes from ScanNet and several in-the-wild scenes. 
Our experimental results show GaussReg's state-of-the-art performance across multiple datasets.
%
%
\bibliographystyle{splncs04}
\bibliography{main}

\begin{thebibliography}{10}
\providecommand{\url}[1]{\texttt{#1}}
\providecommand{\urlprefix}{URL }
\providecommand{\doi}[1]{https://doi.org/#1}

\bibitem{Bai_2020_CVPR}
Bai, X., Luo, Z., Zhou, L., Fu, H., Quan, L., Tai, C.L.: D3feat: Joint learning of dense detection and description of 3d local features. In: Proceedings of the IEEE/CVF Conference on Computer Vision and Pattern Recognition (CVPR) (June 2020)

\bibitem{9710056}
Barron, J.T., Mildenhall, B., Tancik, M., Hedman, P., Martin-Brualla, R., Srinivasan, P.P.: Mip-nerf: A multiscale representation for anti-aliasing neural radiance fields. In: 2021 IEEE/CVF International Conference on Computer Vision (ICCV). pp. 5835--5844 (2021). \doi{10.1109/ICCV48922.2021.00580}

\bibitem{9878829}
Barron, J.T., Mildenhall, B., Verbin, D., Srinivasan, P.P., Hedman, P.: Mip-nerf 360: Unbounded anti-aliased neural radiance fields. In: 2022 IEEE/CVF Conference on Computer Vision and Pattern Recognition (CVPR). pp. 5460--5469 (2022). \doi{10.1109/CVPR52688.2022.00539}

\bibitem{121791}
Besl, P., McKay, N.D.: A method for registration of 3-d shapes. IEEE Transactions on Pattern Analysis and Machine Intelligence  \textbf{14}(2),  239--256 (1992). \doi{10.1109/34.121791}

\bibitem{Chen2022ECCV}
Chen, A., Xu, Z., Geiger, A., Yu, J., Su, H.: Tensorf: Tensorial radiance fields. In: European Conference on Computer Vision (ECCV) (2022)

\bibitem{Chen_2023_ICCV}
Chen, Y., Lee, G.H.: Dreg-nerf: Deep registration for neural radiance fields. In: Proceedings of the IEEE/CVF International Conference on Computer Vision (ICCV). pp. 22703--22713 (October 2023)

\bibitem{dai2017bundlefusion}
Dai, A., Nie{\ss}ner, M., Zoll{\"o}fer, M., Izadi, S., Theobalt, C.: Bundlefusion: Real-time globally consistent 3d reconstruction using on-the-fly surface re-integration. ACM Transactions on Graphics 2017 (TOG)  (2017)

\bibitem{objaverseXL}
Deitke, M., Liu, R., Wallingford, M., Ngo, H., Michel, O., Kusupati, A., Fan, A., Laforte, C., Voleti, V., Gadre, S.Y., VanderBilt, E., Kembhavi, A., Vondrick, C., Gkioxari, G., Ehsani, K., Schmidt, L., Farhadi, A.: Objaverse-xl: A universe of 10m+ 3d objects. arXiv preprint arXiv:2307.05663  (2023)

\bibitem{DeTone_2018_CVPR_Workshops}
DeTone, D., Malisiewicz, T., Rabinovich, A.: Superpoint: Self-supervised interest point detection and description. In: Proceedings of the IEEE Conference on Computer Vision and Pattern Recognition (CVPR) Workshops (June 2018)

\bibitem{fang2023nerfuser}
Fang, J., Lin, S., Vasiljevic, I., Guizilini, V., Ambrus, R., Gaidon, A., Shakhnarovich, G., Walter, M.R.: Nerfuser: Large-scale scene representation by nerf fusion (2023)

\bibitem{5226635}
Furukawa, Y., Ponce, J.: Accurate, dense, and robust multiview stereopsis. IEEE Transactions on Pattern Analysis and Machine Intelligence  \textbf{32}(8),  1362--1376 (2010). \doi{10.1109/TPAMI.2009.161}

\bibitem{1641048}
Goesele, M., Curless, B., Seitz, S.: Multi-view stereo revisited. In: 2006 IEEE Computer Society Conference on Computer Vision and Pattern Recognition (CVPR'06). vol.~2, pp. 2402--2409 (2006). \doi{10.1109/CVPR.2006.199}

\bibitem{gojcic20193DSmoothNet}
Gojcic, Z., Zhou, C., Wegner, J.D., Andreas, W.: The perfect match: 3d point cloud matching with smoothed densities. In: International conference on computer vision and pattern recognition (CVPR) (2019)

\bibitem{goli2023nerf2nerf}
Goli, L., Rebain, D., Sabour, S., Garg, A., Tagliasacchi, A.: nerf2nerf: Pairwise registration of neural radiance fields. In: International Conference on Robotics and Automation (ICRA). IEEE (2023)

\bibitem{HernndezEsteban2004}
Hernández~Esteban, C., Schmitt, F.: Silhouette and stereo fusion for 3d object modeling. Computer Vision and Image Understanding  \textbf{96}(3),  367–392 (Dec 2004). \doi{10.1016/j.cviu.2004.03.016}, \url{http://dx.doi.org/10.1016/j.cviu.2004.03.016}

\bibitem{kerbl3Dgaussians}
Kerbl, B., Kopanas, G., Leimk{\"u}hler, T., Drettakis, G.: 3d gaussian splatting for real-time radiance field rendering. ACM Transactions on Graphics  \textbf{42}(4) (July 2023), \url{https://repo-sam.inria.fr/fungraph/3d-gaussian-splatting/}

\bibitem{9318535}
Li, J., Hu, Q., Ai, M.: Point cloud registration based on one-point ransac and scale-annealing biweight estimation. IEEE Transactions on Geoscience and Remote Sensing  \textbf{59}(11),  9716--9729 (2021). \doi{10.1109/TGRS.2020.3045456}

\bibitem{7349220}
Mellado, N., Dellepiane, M., Scopigno, R.: Relative scale estimation and 3d registration of multi-modal geometry using growing least squares. IEEE Transactions on Visualization and Computer Graphics  \textbf{22}(9),  2160--2173 (2016). \doi{10.1109/TVCG.2015.2505287}

\bibitem{mildenhall2020nerf}
Mildenhall, B., Srinivasan, P.P., Tancik, M., Barron, J.T., Ramamoorthi, R., Ng, R.: Nerf: Representing scenes as neural radiance fields for view synthesis. In: ECCV (2020)

\bibitem{mueller2022instant}
M\"uller, T., Evans, A., Schied, C., Keller, A.: Instant neural graphics primitives with a multiresolution hash encoding. ACM Trans. Graph.  \textbf{41}(4),  102:1--102:15 (Jul 2022). \doi{10.1145/3528223.3530127}, \url{https://doi.org/10.1145/3528223.3530127}

\bibitem{5432191}
Myronenko, A., Song, X.: Point set registration: Coherent point drift. IEEE Transactions on Pattern Analysis and Machine Intelligence  \textbf{32}(12),  2262--2275 (2010). \doi{10.1109/TPAMI.2010.46}

\bibitem{9156303}
Pais, G.D., Ramalingam, S., Govindu, V.M., Nascimento, J.C., Chellappa, R., Miraldo, P.: 3dregnet: A deep neural network for 3d point registration. In: 2020 IEEE/CVF Conference on Computer Vision and Pattern Recognition (CVPR). pp. 7191--7201 (2020). \doi{10.1109/CVPR42600.2020.00722}

\bibitem{8490968}
Pan, Y., Yang, B., Liang, F., Dong, Z.: Iterative global similarity points: A robust coarse-to-fine integration solution for pairwise 3d point cloud registration. In: 2018 International Conference on 3D Vision (3DV). pp. 180--189 (2018). \doi{10.1109/3DV.2018.00030}

\bibitem{Paris2006}
Paris, S., Sillion, F.X., Quan, L.: A surface reconstruction method using global graph cut optimization. International Journal of Computer Vision  \textbf{66}(2),  141--161 (Feb 2006). \doi{10.1007/s11263-005-3953-x}, \url{https://doi.org/10.1007/s11263-005-3953-x}

\bibitem{paszke2017automatic}
Paszke, A., Gross, S., Chintala, S., Chanan, G., Yang, E., DeVito, Z., Lin, Z., Desmaison, A., Antiga, L., Lerer, A.: Automatic differentiation in pytorch  (2017)

\bibitem{9879611}
Qin, Z., Yu, H., Wang, C., Guo, Y., Peng, Y., Xu, K.: Geometric transformer for fast and robust point cloud registration. In: 2022 IEEE/CVF Conference on Computer Vision and Pattern Recognition (CVPR). pp. 11133--11142 (2022). \doi{10.1109/CVPR52688.2022.01086}

\bibitem{sarlin2019coarse}
Sarlin, P.E., Cadena, C., Siegwart, R., Dymczyk, M.: From coarse to fine: Robust hierarchical localization at large scale. In: CVPR (2019)

\bibitem{sarlin20superglue}
Sarlin, P.E., DeTone, D., Malisiewicz, T., Rabinovich, A.: {SuperGlue}: Learning feature matching with graph neural networks. In: CVPR (2020), \url{https://arxiv.org/abs/1911.11763}

\bibitem{7780814}
Schönberger, J.L., Frahm, J.M.: Structure-from-motion revisited. In: 2016 IEEE Conference on Computer Vision and Pattern Recognition (CVPR). pp. 4104--4113 (2016). \doi{10.1109/CVPR.2016.445}

\bibitem{10.1007/978-3-7091-6756-4_6}
Slabaugh, G., Schafer, R., Malzbender, T., Culbertson, B.: A survey of methods for volumetric scene reconstruction from photographs. In: Mueller, K., Kaufman, A.E. (eds.) Volume Graphics 2001. pp. 81--100. Springer Vienna, Vienna (2001)

\bibitem{1641047}
Strecha, C., Fransens, R., Van~Gool, L.: Combined depth and outlier estimation in multi-view stereo. In: 2006 IEEE Computer Society Conference on Computer Vision and Pattern Recognition (CVPR'06). vol.~2, pp. 2394--2401 (2006). \doi{10.1109/CVPR.2006.78}

\bibitem{Sun_2022_CVPR}
Sun, C., Sun, M., Chen, H.T.: Direct voxel grid optimization: Super-fast convergence for radiance fields reconstruction. In: Proceedings of the IEEE/CVF Conference on Computer Vision and Pattern Recognition (CVPR). pp. 5459--5469 (June 2022)

\bibitem{thomas2019KPConv}
Thomas, H., Qi, C.R., Deschaud, J.E., Marcotegui, B., Goulette, F., Guibas, L.J.: Kpconv: Flexible and deformable convolution for point clouds. Proceedings of the IEEE International Conference on Computer Vision  (2019)

\bibitem{wang2023f2nerf}
Wang, P., Liu, Y., Chen, Z., Liu, L., Liu, Z., Komura, T., Theobalt, C., Wang, W.: F2-nerf: Fast neural radiance field training with free camera trajectories. CVPR  (2023)

\bibitem{Wang_2019_ICCV}
Wang, Y., Solomon, J.M.: Deep closest point: Learning representations for point cloud registration. In: Proceedings of the IEEE/CVF International Conference on Computer Vision (ICCV) (October 2019)

\bibitem{wu2023clnerf}
Wu, X., Dai, P., DENG, W., Chen, H., Wu, Y., Cao, Y.P., Shan, Y., QI, X.: {CL}-ne{RF}: Continual learning of neural radiance fields for evolving scene representation. In: Thirty-seventh Conference on Neural Information Processing Systems (2023), \url{https://openreview.net/forum?id=uZjpSBTPik}

\bibitem{10.1007/978-3-030-01237-3_47}
Yao, Y., Luo, Z., Li, S., Fang, T., Quan, L.: Mvsnet: Depth inference for unstructured multi-view stereo. In: Ferrari, V., Hebert, M., Sminchisescu, C., Weiss, Y. (eds.) Computer Vision -- ECCV 2018. pp. 785--801. Springer International Publishing, Cham (2018)

\bibitem{Yew_2022_CVPR}
Yew, Z.J., Lee, G.H.: Regtr: End-to-end point cloud correspondences with transformers. In: Proceedings of the IEEE/CVF Conference on Computer Vision and Pattern Recognition (CVPR). pp. 6677--6686 (June 2022)

\bibitem{10.1007/978-3-540-76390-1_17}
Zaharescu, A., Boyer, E., Horaud, R.: Transformesh : A topology-adaptive mesh-based approach to surface evolution. In: Yagi, Y., Kang, S.B., Kweon, I.S., Zha, H. (eds.) Computer Vision -- ACCV 2007. pp. 166--175. Springer Berlin Heidelberg, Berlin, Heidelberg (2007)

\bibitem{8897021}
Zang, Y., Lindenbergh, R., Yang, B., Guan, H.: Density-adaptive and geometry-aware registration of tls point clouds based on coherent point drift. IEEE Geoscience and Remote Sensing Letters  \textbf{17}(9),  1628--1632 (2020). \doi{10.1109/LGRS.2019.2950128}

\bibitem{zeng20163dmatch}
Zeng, A., Song, S., Nie{\ss}ner, M., Fisher, M., Xiao, J., Funkhouser, T.: 3dmatch: Learning local geometric descriptors from rgb-d reconstructions. In: CVPR (2017)

\bibitem{9336308}
Zhang, J., Yao, Y., Deng, B.: Fast and robust iterative closest point. IEEE Transactions on Pattern Analysis and Machine Intelligence  \textbf{44}(7),  3450--3466 (2022). \doi{10.1109/TPAMI.2021.3054619}

\bibitem{zhang20233d}
Zhang, X., Yang, J., Zhang, S., Zhang, Y.: 3d registration with maximal cliques. In: Proceedings of the IEEE/CVF Conference on Computer Vision and Pattern Recognition. pp. 17745--17754 (2023)

\bibitem{10.1007/978-3-319-46475-6_47}
Zhou, Q.Y., Park, J., Koltun, V.: Fast global registration. In: Leibe, B., Matas, J., Sebe, N., Welling, M. (eds.) Computer Vision -- ECCV 2016. pp. 766--782. Springer International Publishing, Cham (2016)

\end{thebibliography}
\end{document}